\documentclass{article}

\usepackage{microtype}
\usepackage{graphicx}
\usepackage{booktabs} 

\usepackage{hyperref}

\usepackage[accepted]{icml2023}

\usepackage{amsmath}
\usepackage{amssymb}
\usepackage{mathtools}
\usepackage{amsthm}

\usepackage{enumitem}

\usepackage[capitalize,noabbrev]{cleveref}

\theoremstyle{plain}
\newtheorem{theorem}{Theorem}[section]

\newtheorem{lemma}[theorem]{Lemma}
\newtheorem{corollary}[theorem]{Corollary}
\theoremstyle{definition}

\theoremstyle{remark}
\newtheorem{remark}[theorem]{Remark}

\usepackage[disable,textsize=tiny]{todonotes}
\usepackage{multirow}
\usepackage{url}
\usepackage{booktabs}
\usepackage{colortbl}
\usepackage[caption = false]{subfig}

\definecolor{Gray}{gray}{0.85}
\definecolor{lightyellow}{rgb}{1.0, 0.98, 0.65}

\definecolor{dark2green}{rgb}{0.1, 0.65, 0.3}
\definecolor{dark2orange}{rgb}{0.9, 0.4, 0.}
\definecolor{dark2purple}{rgb}{0.4, 0.4, 0.8}
\newcommand{\first}[1]{\textbf{\textcolor{dark2green}{#1}}}
\newcommand{\second}[1]{\textbf{\textcolor{dark2orange}{#1}}}
\newcommand{\third}[1]{\textbf{\textcolor{dark2purple}{#1}}}

\newcolumntype{a}{>{\columncolor{Gray}}c}

\newcommand{\R}{\mathbb{R}}
\newcommand{\Z}{\mathbb{Z}}
\newcommand{\diam}{\mathrm{diam}}

\let\mytodo\todo

\begin{document}

\twocolumn[
\icmltitle{\textsc{Exphormer}: Sparse Transformers for Graphs}



\icmlsetsymbol{equal}{*}

\begin{icmlauthorlist}
\icmlauthor{Hamed Shirzad}{equal,ubc}
\icmlauthor{Ameya Velingker}{equal,google-research}
\icmlauthor{Balaji Venkatachalam}{equal,google}
\icmlauthor{Danica J.\ Sutherland}{ubc,amii}
\icmlauthor{Ali Kemal Sinop}{google-research}
\end{icmlauthorlist}

\icmlaffiliation{ubc}{Department of Computer Science, University of British Columbia, Vancouver, BC, Canada}
\icmlaffiliation{google-research}{Google Research, Mountain View, California, USA}
\icmlaffiliation{google}{Google, Mountain View, California, USA}
\icmlaffiliation{amii}{Alberta Machine Intelligence Institute, Edmonton, Alberta, Canada}

\icmlcorrespondingauthor{Hamed Shirzad}{shirzad@cs.ubc.ca}
\icmlcorrespondingauthor{Ameya Velingker}{ameyav@google.com}
\icmlcorrespondingauthor{Balaji Venkatachalam}{bave@google.com}

\icmlkeywords{Machine Learning, Graph Neural Networks, GNN, Transformer, Attention}

\vskip 0.3in
]



\printAffiliationsAndNotice{\icmlEqualContribution} 

\begin{abstract}
 Graph transformers have emerged as a promising architecture for a variety of graph learning and representation tasks. Despite their successes, though, it remains challenging to scale graph transformers to large graphs while maintaining accuracy competitive with message-passing networks. In this paper, we introduce \textsc{Exphormer}, a framework for building powerful and scalable graph transformers. \textsc{Exphormer} consists of a sparse attention mechanism based on two mechanisms: virtual \emph{global nodes} and \emph{expander graphs}, whose mathematical characteristics, such as spectral expansion, pseduorandomness, and sparsity, yield graph transformers with complexity only linear in the size of the graph, while allowing us to prove desirable theoretical properties of the resulting transformer models. We show that incorporating \textsc{Exphormer} into the recently-proposed GraphGPS framework produces models with competitive empirical results on a wide variety of graph datasets, including state-of-the-art results on three datasets. We also show that \textsc{Exphormer} can scale to datasets on larger graphs than shown in previous graph transformer architectures. Code can be found at \url{https://github.com/hamed1375/Exphormer}.
\end{abstract}

\section{Introduction}
Graph learning has become an important and popular area of study that has yielded impressive results on a wide variety of graphs and tasks, including molecular graphs, social network graphs, knowledge graphs, and more. 
While much research around graph learning has focused on graph neural networks (GNNs), which are based on local \emph{message-passing}, a more recent approach to graph learning that has garnered much interest involves the use of \emph{graph transformers} (GTs). Graph transformers largely operate by encoding graph structure in the form of a \emph{soft inductive bias}. These can be viewed as a graph adaptation of the Transformer architecture~\citep{VaswaniSPUJGKP17} that are successful in modeling sequential data in applications such as natural language processing. 

Graph transformers allow nodes to attend to all other nodes in a graph, allowing for direct modeling of long-range interactions, in contrast to GNNs. This allows them to avoid several limitations associated with local message passing GNNs, such as oversmoothing~\citep{OonoSuzuki20}, oversquashing~\citep{AlonYahav21, ToppingGCDB22}, and limited expressivity~\citep{morris2019weisfeiler, xu2018powerful}. The promise of graph transformers has led to a large number of different graph transformer models that have been proposed in recent years~\citep{DwivediBresson21, kreuzer2021rethinking, Ying2021DoTR, MialonCSM21}.  

One major challenge for graph transformers is their poor scalability, as the standard global attention mechanism incurs time and memory complexity of $O(|V|^2)$, \emph{quadratic} in the number of nodes in the graph. While this cost is often acceptable for datasets with small graphs (e.g., molecular graphs), it can be prohibitively expensive for datasets containing larger graphs, where graph transformer models often do not fit in memory even for high-memory GPUs, and hence would require much more complex and slower schemes to apply.  Moreover, despite the expressivity advantages of graph transformer networks~\citep{kreuzer2021rethinking}, these architectures have often lagged message-passing counterparts in accuracy in many practical settings.

A recent breakthrough came with the advent of GraphGPS~\citep{RampasekGDLWB22}, a modular framework for constructing networks by combining local message passing and a global attention mechanism together with a choice of various positional and structural encodings. To attempt to overcome the quadratic complexity of the ``dense'' full transformer and improve scalability, the architecture allows for ``sparse'' attention mechanisms, like Performer~\citep{ChoromanskiLDSG21} or Big Bird~\citep{ZaheerGDAAOPRWY20}.
This combination of Transformers and GNNs achieves state-of-the-art performance on a wide variety of datasets.

In almost all cases, however, the best results of \citeauthor{RampasekGDLWB22} were obtained by combining a message-passing network with a full transformer;
their sparse transformers performed relatively poorly by comparison,
and indeed their ablation studies showed that on a number of datasets it is better to avoid using attention at all than to use their implementation of BigBird.
This may be related to the fact that sparse attention mechanisms like BigBird have largely been designed for \emph{sequences}; this is natural for language tasks, but graphs behave quite differently.
Thus, it is natural to ask whether one can design sparse attention mechanisms more tailored to learning interactions on general graphs.

Another major question concerns graph transformers' scalability. While BigBird and Performer are linear attention mechanisms, they still incur computational overhead that dominates the per-epoch computation time for moderately-sized graphs. The GraphGPS work tackles datasets with graphs of up to 5,000 nodes, a regime in which the full-attention transformer is in fact computationally faster than many sparse linear-attention mechanisms. Perhaps more suitable sparse attention mechanisms could enable their framework to operate on even larger graphs. Additionally, for smaller graphs, they may require a much smaller batch size to fit into the GPU memory, resulting in slower training, or the need for high-end GPUs. For instance, on the MalNet-Tiny dataset, GraphGPS with a full transformer runs out of memory (on a 40GB NVIDIA A100 GPU) with a batch size of just 16; as we shall see, our techniques allow us to extend the batch size to 256 without any memory issues.

\paragraph{Our contributions.}
We propose sparse attention mechanisms with computational cost linear in the number of nodes and edges. We introduce \textsc{Exphormer} that combines the two techniques for creating sparse overlay graphs.  The first sparse attention mechanism is to use \emph{global  nodes} --- nodes that are connected to all other nodes of the graph. The number of additional edges added to the graph is linear in the number of nodes.
We also introduce \emph{expander graphs} as a powerful primitive in designing scalable graph transformer architectures. The expander graph as an overlay graph has edges linear in the number of nodes.
Expander graphs have several desirable properties -- small diameter, spectral approximation of a complete graph, good mixing properties --  which make them a suitable ingredient in a sparse attention mechanism. We are able to show that \textsc{Exphormer}, which combines expander graphs with global nodes and local neighborhoods, spectrally approximates the full attention mechanism with only a small number of layers, and has universal approximation properties.
\mytodo{Modify the previous section according to the final theoretical results.}
Its attention scheme provides good inductive bias for places the model ``should look,''
in addition to being more efficient and less memory-intensive.

We show that \textsc{Exphormer} are a powerful Transformer that produces results comparable to  GraphGPS with a full transformer\todo{On some datasets in Tables 9-11, or always?}.  Moreover, when combined with MPNNs in the GraphGPS framework, we can achieve SOTA or close to SOTA.
\textsc{Exphormer}
(1) produces better results than other sparse transformers on all datasets we evaluate; 
(2) has results comparable to and in some cases better than the full transformer, despite having fewer parameters;
(3) achieves state-of-the-art results on many datasets, better than MPNNs, including on Long Range Graph Benchmark (LRGB; \citealp{lrgb-2022}) datasets that require long-range dependencies between nodes;
and (4) can scale to larger graphs than previously shown.\mytodo{CIFAR10, MNIST, and PATTERN}

\section{Related Work}
\paragraph{Graph Neural Networks (GNNs).} Early works in the area of graph learning and GNNs include the development of a number of architectures such as GCN~\citep{defferrard2016convolutional, kipf2017semi}, GraphSage~\citep{hamilton2017inductive}, GIN~\citep{xu2018powerful}, GAT~\citep{velickovic2018graph}, GatedGCN~\citep{bresson2017residual}, and more. GNNs are based on a message-passing architecture that generally confines their expressivity to the limits of the  1-Weisfeiler-Lehman (1-WL) isomorphism test~\citep{xu2018powerful}.

A number of recent papers have sought to augment GNNs to improve their expressivity. For instance, one approach has been to use \emph{additional features} that allow nodes to be distinguished -- such as using a {one-hot encoding} of the node~\citep{murphy2019relational} or a {random} scalar feature~\citep{sato2021random} -- or to encode positional or structural information of the graph -- e.g., skip-gram based network embeddings~\citep{qiu2018network}, substructure counts~\citep{bouritsas2020improving}, or Laplacian eigenvectors~\citep{dwivedi2021graph}. Another direction has been to \emph{modify the message passing rule} to allow the network to take further advantage of the graph structure -- including directional graph networks (DGN; \citealp{beaini2021directional}) that use Laplacian eigenvectors to define {directional flows} for anisotropic message aggregation -- or to \emph{modify the underlying graph} over which message passing occurs with \emph{higher-order GNNs}~\citep{morris2019weisfeiler} or the use of substructures such as junction trees~\citep{fey2020hierarchical} and simplicial complexes~\citep{bodnar2021weisfeiler}.%
\footnote{\citet{expander-graph-prop}, in work concurrent to and independent of ours, propose an expander-based graph learning mechanism for message-passing networks, alternating between layers based on the input graph with ones based on an auxiliary expander graph. This scheme is rather different from ours; we do not compare further.}

\paragraph{Graph transformer architectures.} Attention mechanisms have been extremely successful in sequence modeling since the seminal work of~\citet{VaswaniSPUJGKP17}. The GAT architecture~\citep{velickovic2018graph} proposed 
using an attention mechanism to determine how a node aggregates
information from its neighbors; it does not 
use a positional encoding for nodes, limiting its
ability to exploit global structural information.
GraphBert~\citep{zhangGraphBert} uses the graph structure
to determine an encoding of the nodes, but not for the underlying attention mechanism.

Graph transformer models typically operate on a fully-connected graph in which every pair of nodes is 
connected, regardless of the connectivity structure of the original graph. Spectral Attention 
Networks (SAN)~\citep{kreuzer2021rethinking} make use of \emph{two} attention mechanisms, one on the fully-connected graph and one on the original edges of the input graph, while using Laplacian positional encodings for the nodes. Graphormer~\citep{Ying2021DoTR} uses a single dense attention mechanism but adds structural features in the form of centrality and spatial encodings. Meanwhile, GraphiT~\citep{MialonCSM21} incorporates relative positional encodings based on diffusion kernels.

GraphGPS~\citep{RampasekGDLWB22} proposed a general framework for combining message-passing networks with attention mechanisms, while allowing for the mixing and matching of positional and structural embeddings. Specifically, the framework also allows for sparse transformer models like BigBird~\citep{ZaheerGDAAOPRWY20} and Performer~\citep{ChoromanskiLDSG21}.

Moreover, recent works have proposed sampling-based scalable graph transformers, e.g., Gophormer~\citep{gophormer21}, NAGphormer~\citep{nagphormer22}. Another line of work aims to learn data dependencies beyond the given graph structure using linear time transformers. For instance, Nodeformer~\citep{wu2022nodeformer} is inspired by Performer~\citep{ChoromanskiLDSG21} and leverages the kernelized Gumbel-Softmax operator to enable the propagation of information among all pairs of nodes in a computationally efficient manner. Another work is Difformer~\citep{wu2023difformer}, which, on the other hand, is a continuous time diffusion-based Transformer model. Even though these models can scale up to millions of nodes, both of them use a random subset of a maximum of 100K nodes as mini-batches, and the attention mechanism takes place only over the nodes in the batch. Transformers are also applied on the spectral GNNs~\citep{bo2023specformer}.

\paragraph{Sparse Transformers.} Standard (dense) transformers have quadratic complexity in the number of tokens, which limits their scalability to extremely long sequences. By contrast, \emph{sparse transformer} models improve computational and memory efficiency by restricting the attention pattern, i.e., the pairs of nodes that can interact with each other.
In addition to BigBird and Performer, there have been a number of other proposals for sparse transformers; \citet{TayDBM20} provide a survey.

\section{Sparse Attention on Graphs}
This section describes three \emph{sparse} patterns that can be used in transformers in individual layers of a graph transformer architecture.
We begin by describing graph attention mechanisms in general.

\subsection{Attention mechanism on graphs}
An attention mechanism on $n$ tokens can be modeled by a directed graph $H$ on $[n] = \{1,2,\dots,n\}$, where 
a directed edge from $i$ to $j$ indicates a direct interaction between tokens $i$ and $j$, i.e., an inner 
product that will be computed by the attention mechanism.
More precisely, a transformer block can be viewed as a function on the $d$-dimensional embeddings for each of $n$ tokens, mapping from $\R^{d\times n}$ to $\R^{d\times n}$. Let $\mathbf{X} = (\mathbf{x}_1, \mathbf{x}_2, \dots, \mathbf{x}_n) \in \R^{d\times n}$. A generalized (dot-product) attention mechanism $\textsc{Attn}_H: \R^{d\times n}\to\R^{d\times n}$ with attention pattern given by $H$ is defined by
\begin{multline*}
 \textsc{Attn}_H(\mathbf{X})_{:,i} = \mathbf{x}_i +  \\
 \sum_{j=1}^h \mathbf{W}_O^j \mathbf{W}_V^j \mathbf{X}_{\mathcal{N}_H(i)} \cdot \sigma\left( \left(\mathbf{W}_K^j \mathbf{X}_{\mathcal{N}_H(i)} \right)^T \left(\mathbf{W}_Q^j \mathbf{x}_i\right) \right),
\end{multline*}
where $h$ is the number of heads and $m$ is the head size, while $\mathbf{W}_K^j, \mathbf{W}_Q^j, \mathbf{W}_V^j \in \R^{m\times d}$ and $\mathbf{W}_O^j \in \R^{d\times m}$.
(The subscript $K$ is for ``keys,'' $Q$ for ``queries,'' $V$ for ``values,'' and $O$ for ``output.'')
Here $\mathbf{X}_{\mathcal{N}_H(i)}$ denotes the submatrix of $\mathbf{X}$ obtained by picking out only those columns corresponding to elements of $\mathcal{N}_H(i)$, the neighbors of $i$ in $H$. We can see that the total number of inner product computations for all $i\in [n]$ is given by the number of edges of $H$. A (generalized) \emph{transformer block} consists of $\textsc{Attn}_H$ followed by a feedforward layer:
\begin{multline*}
 \textsc{FF}(\mathbf{X}) = \textsc{Attn}_H(\mathbf{X}) \\+ 
 \mathbf{W}_2 \cdot \text{ReLU}(\mathbf{W}_1 \cdot \textsc{Attn}_H\left(\mathbf{X}) + \mathbf{b}_1 \mathbf{1}_n^T\right) + \mathbf{b}_2 \mathbf{1}_n^T,
\end{multline*}
where $\mathbf{W}_1 \in \R^{r\times d}$, $\mathbf{W}_2 \in \R^{d\times r}$, $\mathbf{b}_1 \in \R^r$, and $\mathbf{b}_2 \in \R^d$.

In the standard setting, the $n$ tokens are part of a sequence (e.g., language applications). However, we are concerned with the \emph{graph transformer} setting in which the tokens are nodes of some underlying graph $G = (V,E)$ with $V=[n]$. The attention computation is nearly identical, except that one can also optionally augment it  with edge features, as is done in SAN~\citep{kreuzer2021rethinking}:
\begin{multline*}
 \textsc{Attn}_H(\mathbf{X})_{:,i} = \mathbf{x}_i + \sum_{j=1}^h \mathbf{W}_O^j \mathbf{W}_V^j \mathbf{X}_{\mathcal{N}_H(i)} \cdot \\
 \sigma\left( \left(\mathbf{W}_E^j \mathbf{E}_{\mathcal{N}_H(i)}  \odot \mathbf{W}_K^j \mathbf{X}_{\mathcal{N}_H(i)} \right)^T \left(\mathbf{W}_Q^j \mathbf{x}_i\right) \right),
\end{multline*}
where $\mathbf{W}_E^j \in \R^{m\times d_E}$, $\mathbf{E}_{\mathcal{N}_H(i)} $ is the $d_E \times |\mathcal{N}_H(i)|$ matrix whose columns are $d_E$-dimensional edge features for the edges connected to node $i$, and $\odot$ denotes element-wise multiplications.

The most typical cases of graph transformers use full (dense) attention, where every token attends to every other node: $H$ is the fully-connected directed graph. As this results in computational complexity $O(n^2)$ for the transformer block, which is prohibitively expensive for large graphs, we wish to replace full attention with a \emph{sparse} attention mechanism, where $H$ has $o(n^2)$ edges -- ideally, $O(n)$. 

A number of sparse attention mechanisms have been proposed to address the aforementioned issue (see~\citealp{TayDBM20}), but the vast majority are designed specifically for functions on \emph{sequences}. \textsc{Exphormer}, on the other hand, is a \emph{graph-centric} sparse attention mechanism that makes use of the underlying structure of the input graph $G$. We introduce three sparse patterns: expander graphs, global connectors, and local neighborhoods as three patterns that can be used in transformers. They can combine together to make an \textsc{Exphormer} layer, but it is not necessary to have all components in each layer. 

GraphGPS~\citep{RampasekGDLWB22} shows an effective way to include the transformer layers beside an MPNN model. We use the same architecture as GraphGPS, replacing only the transformer part with an \textsc{Exphormer} layer. We will show that sparse transformers can get comparable and even better results on many benchmarks compared to full transformers, or sparse transformer variants originally designed for sequences.

\subsection{The \textsc{Exphormer} Architecture}\label{sec:exphormerarch}

\begin{figure}[bt]
\vspace{-0.1in}
\captionsetup[subfloat]{}
\centering
\subfloat[][]{\includegraphics[width = 0.8in]{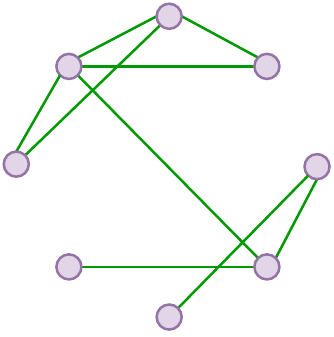}} 
\hspace{0.3in}
\subfloat[][]{\includegraphics[width = 0.8in]{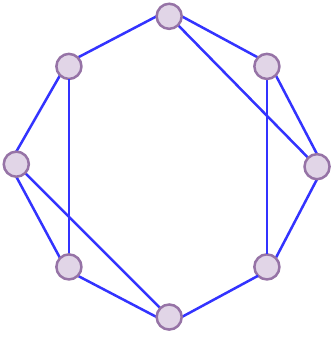}}
\\
\subfloat[][]{\includegraphics[width = 0.8in]{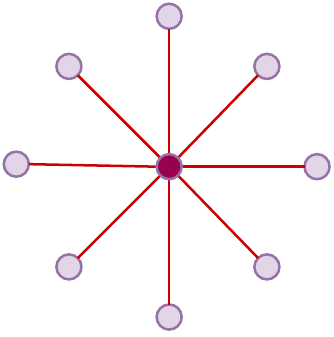}}
\hspace{0.3in}
\subfloat[][]{\includegraphics[width = 0.8in]{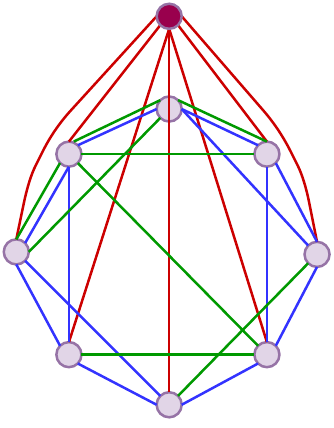}}

\caption{The components of \textsc{Exphormer}: (a) shows local neighborhood attention, i.e., edges of the input graph. (b) shows an expander graph with degree 3. (c) shows global attention with a single virtual node. (d) All of the aforementioned components are combined into a single interaction graph that determines the attention pattern of \textsc{Exphormer}.}
\label{fig:exphormer}
\vspace{-0.1in}
\end{figure}

\begin{table*}[t]
    \vspace{-0.1in}
    \centering
    \caption{Comparison of \textsc{Exphormer} with baselines on various datasets. Best results are colored: \first{first}, \second{second}, \third{third}.}
    \label{tab:benchmarking}
    \fontsize{8.25pt}{8.25pt}\selectfont
    \setlength\tabcolsep{6.25pt} 
    \scalebox{0.9}{
    \begin{tabular}{p{4cm}ccccc}
    \toprule
         {\bf Model} & {\bf CIFAR10} & {\bf MalNet-Tiny} & {\bf MNIST} & {\bf CLUSTER} & {\bf PATTERN} \\
          & {Accuracy $\uparrow$} & {Accuracy $\uparrow$} & {Accuracy $\uparrow$} & {Accuracy $\uparrow$} & {Accuracy $\uparrow$}\\
    \midrule
    GCN {\tiny\cite{kipf2017semi}} & 55.71$\pm$0.381 & 81.0 & 90.71$\pm$0.218 &  68.50 $\pm$ 0.976 & 71.89 $\pm$ 0.334\\
    GIN {\tiny\cite{xu2018powerful}} & 55.26$\pm$1.527 & 88.98$\pm$0.557 & 96.49$\pm$0.252 &  64.72 $\pm$ 1.553 & 85.39 $\pm$ 0.136\\
    GAT {\tiny\cite{velickovic2018graph}} & 64.22$\pm$0.455 & 92.1 $\pm$0.242 & 95.54$\pm$0.205 & 70.59 $\pm$ 0.447 &  78.27 $\pm$ 0.186\\
    GatedGCN {\tiny\cite{bresson2017residual,DwivediJLBB20}} \hangindent=1em & 67.31$\pm$0.311 & \third{92.23$\pm$0.65} & 97.34$\pm$0.143 & 73.84 $\pm$ 0.326 & 85.57 $\pm$ 0.088 \\
    PNA {\tiny\cite{corso2020principal}} & 70.35$\pm$0.63 & -- & 97.94$\pm$0.12 & -- & -- \\
    DGN {\tiny\cite{beaini2021directional}} & \third{72.84$\pm$0.417} & -- & -- & -- & 86.68$\pm$0.034   \\
    \midrule
    CRaWl~{\tiny\cite{toenshoff2021CRaWl}} & 69.01$\pm$0.259 & -- &  97.94$\pm$0.050 & -- & -- \\
    GIN-AK+~{\tiny\cite{zhao2021stars}} & 72.19$\pm$0.13 & -- & -- & -- & \first{86.85$\pm$0.057}  \\
    \midrule
    SAN {\tiny \cite{kreuzer2021rethinking}} & -- & -- & -- & 76.69$\pm$0.65  & 86.58$\pm$0.037 \\
    K-Subgraph SAT {\tiny \cite{chen2022SAT}} & -- & -- & -- & 77.86$\pm$0.104  & \second{86.85$\pm$0.037} \\
    EGT {\tiny \cite{hussain2021edge}} & 68.70$\pm$0.409 & & \second{98.17$\pm$0.087} & \first{79.23$\pm$0.348} & \third{86.82$\pm$0.020} \\
    GraphGPS {\tiny\cite{RampasekGDLWB22}} &  \third{72.30$\pm$0.356} & \second{93.50$\pm$0.41} &  \third{98.05$\pm$0.126} &   \third{78.02$\pm$0.180}  &  86.69$\pm$0.059 \\
    \midrule
    \rowcolor{lightyellow} \textsc{Exphormer} (ours) & \first{74.69$\pm$0.125} & \first{	94.02 $\pm$ 0.209} & \first{98.55 $\pm$ 0.039} & \second{78.07 $\pm$ 0.037} & 86.74$\pm$0.015 \\
     \bottomrule
    \end{tabular}
    }

\end{table*}

We now describe the details of the construction of \textsc{Exphormer}, an expander-based sparse attention mechanism for graph transformers with $O(|V|+|E|)$ computation, where $G=(V,E)$ is the underlying input graph. The \textsc{Exphormer} architecture constructs an interaction graph $H$ that consists of three main components, as shown in \Cref{fig:exphormer}. The construction always has bidirectional edges, and so $H$ can be viewed as an undirected graph.
The mechanism uses three types of edges:
\begin{enumerate}
    \item {\bf Expander graph attention}: Edges from a random \emph{expander graph} can be used as attention patterns. These graphs have several useful theoretical properties related to spectral approximation and random walk mixing (see~\cref{sec:theoreticalprop}), which allow propagating information between pairs of nodes that are distant in the input graph $G$ without connecting all pairs of nodes. They introduce many alternative short paths between the nodes and avoid the information bottleneck that can be caused by the virtual nodes. In particular, we use a regular expander graph of constant degree, which allows the number of edges to be just $O(|V|)$. The process we use to construct a random expander graph is described below.
    
    \item {\bf Global attention}: The next component is \emph{global attention}, whereby a small number of virtual nodes are added to the interaction graph, and each such node is connected to all the non-virtual nodes. These nodes enable a global ``storage sink'' and they have universal approximator functions for full transformers. We will generally add a constant number of virtual nodes, in which case the total number of edges due to global attention will be $O(|V|)$.
    \item {\bf Local neighborhood attention}: Another desirable property to capture is \emph{locality}. Graphs carry much more topological structure than sequences, and the neighborhoods of individual nodes carry a lot of information about connectivity.
    Thus, we model local interactions by allowing each node $v$ to attend to every other node that is an immediate neighbor of $v$ in $G$:
    that is, $H$ includes the input graph edges $E$ as well as their reverses, introducing $O(|E|)$ interaction edges.
    One generalization would be to allow direct attention within $k$-hop neighborhoods, but this might introduce a superlinear number of interactions on general graphs.

\end{enumerate}

We use learnable embeddings for expander and global connection edge features, and virtual nodes' features. Dataset edge features are used for the local neighborhood edge features.
We always use local neighborhoods in \textsc{Exphormer} layers, but expander graphs and global attention are not always helpful; depending on the dataset, we may use both, or only one or the other.
We discuss this further in \cref{sec:expander-vs-global}.

Some previous graph-oriented transformers,
such as the SAN architecture \citep{kreuzer2021rethinking},
use separate attention mechanisms for different sources of edges.
By using a single attention mechanism,
\textsc{Exphormer} achieves a more compact model
that can still distinguish the ``types'' of attention edges based on edge features.

\paragraph{Generating a Random Regular Expander}
We now describe how we generate a random regular expander. Let $G = (V,E)$ be the original graph, where $V = \{1,2,\dots, n\}$. For the purposes of experimentation (in \cref{tab:benchmarking,tab:sparse-comparision,tab:lrgb,tab:largegraphs,tab:alations_comps}), we use the random graph process analyzed in \citet{Friedman03} (see \cref{thm:randomregular} in \cref{sec:expander-details}) to generate a random $d$-regular graph $G' = (V, E')$ on the same node set $V$:
\begin{itemize}[itemsep=2pt,parsep=2pt]
    \item Pick $d/2$ permutations $\pi_1, \pi_2, \dots, \pi_{d/2}$ on $V$, each $\pi_i$ chosen independently and uniformly among all possible permutations of $n$ elements.
    \item Then, letting $[k]$ denote $\{1, 2, \dots, k\}$, choose
    \[ E' = \left\{ (i, \pi_j(i)), (i, \pi_j^{-1}(i)): j \in [d/2], i \in [n]\right\}.\]
\end{itemize}
The above process works for even $d$, and \cref{thm:randomregular} shows that the resulting graph will be a $d$-regular near-Ramanujan graph with high probability. In practice, we will generate expander graphs using this process and throw out any graphs that fail to be near-Ramanujan, according to a desired threshold (this is a low probability event).

We provide further details in \cref{sec:expander-details}, where we also discuss other algorithms for generating expander graphs.


\section{Theoretical Properties of \textsc{Exphormer}} \label{sec:theoreticalprop}
\textsc{Exphormer} is based on expander graphs, which have a number of properties that make them suitable as a key building block of our approach. In this section, we describe relevant properties of expander graphs along with their implications for \textsc{Exphormer}s.

\subsection{Expander Graphs Approximate Complete Graphs}
For simplicity, let us consider $d$-\emph{regular} graphs (where every node has  $d$ neighbors). Suppose $G$ is a $d$-regular undirected graph on $n$ vertices. Let $A_G$ be the $n\times n$ adjacency matrix of $G$. It is known that $A_G$ has $n$ real eigenvalues $d = \lambda_1 \geq \lambda_2 \geq \cdots \geq \lambda_n \geq -d$. The graph $G$ is said to be an \emph{$\epsilon$-expander} if $\max\{|\lambda_2|, |\lambda_n|\} \leq \epsilon d$ \citep{hoory06}.

Expander graphs are sparse approximations of complete graphs.
As we discuss,
expander graphs with only $O(n)$ edges
exist that can preserve certain desirable properties of the complete graph with $\Theta(n^2)$ edges.

\subsubsection{Spectral Properties}
A useful tool to study expanders is the \emph{Laplacian} matrix of a graph, which captures several important spectral properties.
Letting $D_G$ denote the $n\times n$ diagonal matrix whose $i$-th diagonal entry is the degree of the $i$-th node, we define $L_G = D_G-A_G$ to be the Laplacian of $G$.
The following theorem is well-known in spectral graph theory.
\begin{theorem} 
\citep[Section 27.2]{spielman:sagt}
 A $d$-regular $\epsilon$-expander $G$ on $n$ vertices spectrally approximates the complete graph $K_n$ on $n$ vertices:\footnote{For matrices $A$ and $B$, we say that $A \preceq B$ if $B-A$ is a positive semi-definite matrix.}
 \[
  (1-\epsilon) \frac{1}{n} L_K \preceq \frac{1}{d} L_G \preceq (1+\epsilon) \frac{1}{n} L_K.
 \]
\end{theorem}
Spectral approximation is known to preserve the cut structure in graphs. As a result, a sparse attention mechanism based on expander edges retains spectral properties of the full attention mechanism: cuts, vertex expansion, and so on.

\subsubsection{Mixing Properties}
Another property of expanders is that random walks mix well.
Let $G = (V, E)$ be a $d$-regular $\epsilon$-expander. 
Consider a random walk $v_0, v_1, v_2, \dots$ on $G$, where
$v_0 \sim \pi^{(0)}$,
and then each subsequent $v_{t+1}$ is one of the $d$ neighbors of $v_t$ chosen uniformly at random.
We then have $v_t \sim \pi^{(t)}$, given recursively by $\pi^{(t+1)} = D_G^{-1} A_G \pi^{(t)}$. 
It turns out that after a logarithmic number of steps, a random walk from a starting probability distribution on the vertices 
is close to uniformly distributed along all nodes of the graph.

\begin{lemma}\label{lem:mixing}
\citep[Theorem 3.2]{hoory06}
Let $G = (V,E)$ be a $d$-regular $\epsilon$-expander graph on $n=|V|$ nodes. For any initial distribution $\pi^{(0)}:V \to \R^+$ and any $\delta > 0$, $\pi^{(t)}$ satisfies
\[
 \|\pi^{(t)} - \tfrac1n\|_1 \le \delta
\]
as long as
$t
\ge \frac{1}{2(1 - \varepsilon)} \log(n / \delta^2)$.
\end{lemma}

In an attention mechanism of a transformer, one can consider the graph of pairwise interactions (i.e., $i$ is connected to $j$ if $i$ and $j$ attend to each other). If the attention mechanism is dense, then each node is connected to every other node and it is trivial to see that every pair of nodes interacts with each other in a single transformer layer. In a \emph{sparse} attention mechanism, on the other hand, some pairs of nodes are not directly connected, meaning that a single transformer layer will not model interactions between all pairs of nodes. However, if we stack transformer layers on top of each other, the stack will be able to model longer range interactions.  In particular, a consequence of the above lemma is that if our sparse attention mechanism is modeled after an $\epsilon$-expander graph, then stacking at least $t = \frac{1}{2(1-\epsilon)}\log(n/\delta^2)$ layers will model ``most'' pairwise interactions between nodes.

Relatedly, the {diameter} of expander graphs is asymptotically logarithmic in the number of nodes.
\begin{theorem}\citep[Section 2.4]{hoory06}
Suppose $G = (V,E)$ is a $d$-regular $\epsilon$-expander graph on $n$ vertices. Then, for every vertex $v$ and $k \geq 0$, the $k$-hop neighborhood $B(v,r) = \{w \in V: \operatorname{dist}(v,w)\leq k\}$ has
 \[
   |B(v,r)| \geq \min\{(1+c)^k, n\}
 \]
 for some constant $ c > 0$ depending on 
 $d, \epsilon$. 
 In particular, we have that $\diam(G) = O_{d,\epsilon}(\log n)$.
\end{theorem}
As a consequence, we obtain the following, which shows that using logarithmically many successive transformer layers allows each node to propagate information to every node.
\begin{corollary}\label{cor:logdiam}
If a sparse attention mechanism on $n$ nodes is a $d$-regular $\epsilon$-expander graph, then stacking $O_{d,\epsilon}(\log n)$ transformer layers models all pairwise node interactions.
\end{corollary}

\subsection{Universal Approximability of \textsc{Exphormer} Models}
The expander graph attention and global attention components of \textsc{Exphormer} ensure that a sublinear number of graph transformer layers is sufficient to allow each node to interact (directly or indirectly) with every other node, alleviating potential underreaching issues (even in the absence of global nodes, $O(\log n)$ layers are sufficient, by \Cref{cor:logdiam}). Nevertheless, a natural question is whether a sparse transformer model based on \textsc{Exphormer} allows universal approximation of functions.

Dense transformer models are known to be universal approximators, i.e., with the use of positional encodings, they can approximate any continuous sequence-to-sequence function on a compact domain arbitrarily closely~\citep{YunBRRK20}. In the realm of graph learning, this provides a compelling motivation for considering graph transformers over standard MPNNs, which are known to be limited in expressive power by the Weisfeiler-Lehman (WL) hierarchy.

Universal approximation properties for dense transformers do not automatically hold for sparse transformers, but we show every continuous function $f: [0,1]^{d\times |V|} \to \R^{d\times |V|}$ can be approximated to any desired accuracy by an \textsc{Exphormer} network using either global attention or a suitable version of expander attention.  Details are in \cref{sec:proofs}.

\begin{table*}[bt]
\caption{Comparison of attention mechanisms in GPS. \textsc{Exphormer} outperforms other {sparse} transformer architectures (BigBird and Performer) while also beating the full transformer GPS models on three of four datasets. Best results are colored in \first{first}, \second{second}, \third{third}.}
\label{tab:sparse-comparision}
\centering
\fontsize{8.25pt}{8.25pt}\selectfont
\setlength\tabcolsep{6.25pt} 
\scalebox{1.0}{
\begin{tabular}{l|cccc} 
\toprule
Model/Dataset   & \textbf{Cifar10}       & \textbf{MalNet-Tiny}   & \textbf{PascalVOC-SP}    & \textbf{Peptides-Func}   \\ 
 & Accuracy $\uparrow$ & Accuracy $\uparrow$ & F1 score $\uparrow$ & AP $\uparrow$ \\
[0.1cm]
\midrule
GPS (MPNN-only)            & 69.948 $\pm$ 0.499 & 92.23 $\pm$ 0.65  & 0.3016 $\pm$ 0.0031 & 0.6159 $\pm$ 0.0048  \\ 
[0.05cm]
\midrule
GPS-BigBird     & 70.480 $\pm$ 0.106 & 92.34 $\pm$ 0.34  & 0.2762 $\pm$ 0.0069 & 0.5854 $\pm$ 0.0079 \\
[0.05cm]
GPS-Performer   & \third{70.670 $\pm$ 0.338} & \third{92.64 $\pm$ 0.78}  & \third{0.3724 $\pm$ 0.0131} & \third{0.6475 $\pm$ 0.0056} \\
[0.05cm]
GPS-Transformer & \second{72.305 $\pm$ 0.344} & \second{93.50 $\pm$ 0.41}  & \second{0.3736 $\pm$ 0.0158} & \first{0.6535 $\pm$ 0.0041} \\
[0.05cm]
\midrule
\textsc{Exphormer}       & \first{74.69$\pm$0.125}   & \first{94.02 $\pm$ 0.21} & \first{0.3975 $\pm$ 0.0037} & \second{0.6527 $\pm$ 0.0043}  \\
\bottomrule
\end{tabular}
}
\end{table*}

\section{Experiments} \label{sec:experiments}
In this section, we evaluate the empirical performance of graph transformer models based on \textsc{Exphormer} on a wide variety of graph datasets with graph prediction and node prediction tasks~\cite{DwivediJLBB20,HuFZDRLCL20,FreitasDNC21,amazon_dataset,namata:mlg12-wkshp}. In particular, we present experiments on fifteen benchmark datasets, including image-based graph datasets (CIFAR10, MNIST, PascalVOC-SP, COCO-SP), 
synthetic SBM datasets (PATTERN, CLUSTER), code graph datasets (MalNet-Tiny), and molecular datasets (Peptides-Func, Peptides-Struct, PCQM-Contact). Moreover, we demonstrate the ability of \textsc{Exphormer} to allow graph transformers to scale to larger graphs (with more than 5,000 nodes) by including results on five transductive graph datasets consisting of citation networks (CS, Physics, ogbn-arxiv) and co-purchasing networks (Computer, Photo).

For the experimental setup, we combine \textsc{Exphormer} together with MPNNs in the GraphGPS framework~\citep{RampasekGDLWB22}, which constructs graph transformer models by composing attention mechanisms with message-passing schemes, together with an appropriate choice of positional and structural encodings.

In addition to comparisons with baselines on the aforementioned datasets, we also run ablation studies on the various attention components of \textsc{Exphormer} (from \cref{sec:exphormerarch}).

In summary, our experiments show that: (a) \textsc{Exphormer} achieves SOTA performance on a variety of datasets, (b) \textsc{Exphormer} consistently outperforms other sparse attention mechanisms while often surpassing dense transformers using fewer parameters, and (c) \textsc{Exphormer} successfully allows GraphGPS to overcome memory bottlenecks and scale to larger graphs (on $> 10,000$ nodes) while still providing competitive performance.

\begin{table*}[t]
    \caption{Comparison of \textsc{Exphormer} with baselines from the Long-Range Graph Benchmarks (LRGB, \citealp{lrgb-2022}). Best results are colored in \first{first}, \second{second}, \third{third}.}
    \label{tab:lrgb}
    \fontsize{8.5pt}{8.5pt}\selectfont
    \setlength\tabcolsep{4pt} 
    \centering
    \begin{tabular}{lccccc}\toprule
    \multirow{2}{*}{\textbf{Model}} &\textbf{PascalVOC-SP} &\textbf{COCO-SP} &\textbf{Peptides-Func} &\textbf{Peptides-Struct} &\textbf{PCQM-Contact} \\
    &F1 score $\uparrow$ &F1 score $\uparrow$ &AP $\uparrow$ &MAE $\downarrow$ &MRR $\uparrow$ \\\midrule
    GCN &0.1268 $\pm$ 0.0060 &0.0841 $\pm$ 0.0010 &0.5930 $\pm$ 0.0023 &0.3496 $\pm$ 0.0013 &0.3234 $\pm$ 0.0006 \\
    GINE &0.1265 $\pm$ 0.0076 &0.1339 $\pm$ 0.0044 &0.5498 $\pm$ 0.0079 &0.3547 $\pm$ 0.0045 &0.3180 $\pm$ 0.0027 \\
    GatedGCN &0.2873 $\pm$ 0.0219 &\third{0.2641 $\pm$ 0.0045} &0.5864 $\pm$ 0.0077 &0.3420 $\pm$ 0.0013 &0.3218 $\pm$ 0.0011 \\
    GatedGCN+RWSE &0.2860 $\pm$ 0.0085 &0.2574 $\pm$ 0.0034 &0.6069 $\pm$ 0.0035 &0.3357 $\pm$ 0.0006 &0.3242 $\pm$ 0.0008 \\ \midrule
    Transformer+LapPE &0.2694 $\pm$ 0.0098 & 0.2618 $\pm$ 0.0031 &0.6326 $\pm$ 0.0126 &\third{0.2529 $\pm$ 0.0016} &0.3174 $\pm$ 0.0020 \\
    SAN+LapPE &\third{0.3230 $\pm$ 0.0039} &\phantom{*}0.2592 $\pm$ 0.0158* &0.6384 $\pm$ 0.0121 &0.2683 $\pm$ 0.0043 &\second{0.3350 $\pm$ 0.0003} \\
    SAN+RWSE & 0.3216 $\pm$ 0.0027 &\phantom{*}0.2434 $\pm$ 0.0156* &\third{0.6439 $\pm$ 0.0075} &0.2545 $\pm$ 0.0012 &\third{0.3341 $\pm$ 0.0006} \\ 
    GraphGPS &\second{0.3748 $\pm$ 0.0109} &\second{0.3412 $\pm$ 0.0044} &\first{0.6535 $\pm$ 0.0041} &\second{0.2500 $\pm$ 0.0005} &0.3337 $\pm$ 0.0006 \\ \midrule
    Exphormer (ours) & \first{0.3975 $\pm$ 0.0037} & \first{0.3455 $\pm$ 0.0009} & 	\second{0.6527 $\pm$ 0.0043} & \first{0.2481 $\pm$ 0.0007} & \first{	0.3637 $\pm$ 0.0020} \\
    \bottomrule
    \end{tabular}
\end{table*}

\subsection{Comparison to Sparse Attention Mechanisms}\label{sec:attentioncomp}
We first discuss a comparison of our \textsc{Exphormer}-based models with other \emph{sparse} transformer architectures, which are a natural first point of comparison. In particular, we perform a series of experiments that compare \textsc{Exphormer}-based architectures with sparse transformer baselines in the GraphGPS framework. More specifically, for each dataset, we compare our best \textsc{Exphormer} model with GraphGPS models that use BigBird and Performer for the underlying attention mechanism.

The results are shown in \Cref{tab:sparse-comparision}. Note that on each of the four highlighted datasets, an \textsc{Exphormer} model outperforms the sparse attention models GPS-BigBird and GPS-Performer. Furthermore, it even outperforms the full (dense) transformer model (GPS-Transformer) on three of the datasets while performing competitively on the fourth. Since GraphGPS models make use of both attention (transformer) and message passing components, we additionally point out that the \textsc{Exphormer}-based sparse attention component is, indeed, providing lift over a standard MPNN baseline that does not make use of attention at all.

\subsection{Benchmarking GNNs Datasets} \label{sec:baselines}
We have showed in \Cref{sec:attentioncomp} that \textsc{Exphormer} consistently outperforms relevant sparse attention baselines. The natural next question is how \textsc{Exphormer} performs relative to a wider range of baselines, including not just graph transformer models but also other architectures such as MPNNs.

\Cref{tab:benchmarking} shows results on five datasets, including four from the Benchmarking GNNs collection~\citep{DwivediJLBB20} as well as the code graph dataset MalNet-Tiny~\citep{FreitasDNC21}.  We note that an \textsc{Exphormer}-based graph transformer with message-passing in the GraphGPS framework yields \emph{state-of-the-art (SOTA) performance on three of the datasets} and is competitive with the best single model accuracies on the remaining datasets.

Observe that on all five datasets, our \textsc{Exphormer} models provide better accuracy than GraphGPS models based on \emph{full (dense) transformers}. Additionally, the \textsc{Exphormer} models outperform the full transformer-based SAN model as well as a variety of MPNN baselines.

\begin{remark}
Our \textsc{Exphormer} models often outperform full transformer models with a much smaller number of parameters. For instance, on PATTERN, our results reported in \Cref{tab:benchmarking} are obtained from an \textsc{Exphormer} model with just 90,000 parameters, compared to 340,000 parameters in the comparable full transformer GraphGPS model! Similarly, on CLUSTER, our \textsc{Exphormer} model uses just 280,000 parameters, as compared to 500,000 parameters for the full transformer GraphGPS model. This results in simpler models with time and memory advantages. For further details on hyperparameters, see \Cref{tab:hyperparams-gps-1} in the appendix.
\end{remark}

\begin{remark}\label{rmk:batchsize}
\textsc{Exphormer} models often enable larger batch sizes during training. On MalNet-Tiny, which contains some graphs with around 5,000 nodes, the full transformer GraphGPS model runs out of memory on a 40GB NVIDIA A100 GPU when using a batch size of just 16, while our sparse \textsc{Exphormer} models handle a batch size of 256.
\end{remark}
\begin{table*}[!t]
    \centering
    \caption{Accuracy of models with different attention mechanisms on transductive graph datasets (numbers in top rows, other than arXiv, are from \citealp{nagphormer22}).  \citeauthor{nagphormer22} did not report NAGphormer results on this dataset.}
    \label{tab:largegraphs}
    \setlength\tabcolsep{7pt} 
    \scalebox{0.75}{
    \begin{tabular}{lccccc}
    \toprule
         {\bf Model} & {\bf ogbn-arxiv} & {\bf Computer}  & {\bf Photo} & {\bf CS}  & {\bf Physics}\\
         \midrule
         SAN & OOM & 89.83 $\pm$ 0.16 & 94.86 $\pm$ 0.10 & 94.51 $\pm$ 0.15 & OOM \\
         GraphGPS &  OOM & OOM & 95.06 $\pm$ 0.13& 93.93 $\pm$ 0.12 & OOM \\
         NAGphormer & NA & 91.22 $\pm$ 0.14 & 95.49 $\pm$ 0.11 & 95.75 $\pm$ 0.09 & 97.34 $\pm$ 0.03 \\
         \midrule
         \textsc{Exphormer} & 72.44 $\pm$ 0.28 & 91.59 $\pm$ 0.31& 	95.27 $\pm$ 0.42 & 95.77 $\pm$ 0.15 & 97.16 $\pm$ 0.13 \\
 \bottomrule
    \end{tabular}
    }
\end{table*}
\begin{table*}[ht!]
\vspace{-0.1in}
\centering
\caption{Results for the full model versus removing each of the components. For the Peptides-Struct dataset, lower is better; for all the others, higher is better.}
\smallskip
\label{tab:alations_comps}
\setlength{\tabcolsep}{5pt}
\renewcommand{\arraystretch}{1.5}
\scalebox{0.75}{
\begin{tabular}{l|l|l|l|l} 
\toprule
Dataset & No Local Edges & No Expander Edges & No Global Nodes & All Components \\ 
\hline
Cifar10  
 & $74.62 \pm0.12$ & $74.53 \pm 0.19$  & $74.68 \pm 0.19$ &$\bf{74.69 \pm 0.13}$   \\ 
\hline
Malnet-Tiny 
 & $92.64 \pm 0.55$ & $\bf{94.02 \pm 0.21}$ & $92.48 \pm 0.33$ & $92.06 \pm 0.18$  \\
 \hline
Pattern & $86.59 \pm 0.03$ & $86.70 \pm 0.02$ & $86.14 \pm 0.08$ & $\bf{86.74 \pm 0.02}$\\
\hline
PascalVOC-SP & $0.3708 \pm 0.0039$ & $0.3588 \pm 0.0013$ & $\bf{0.3975 \pm 0.0037}$ & $0.3682 \pm 0.0042$\\
\hline
Peptides-Struct & $0.2631 \pm 0.0007$ & $\bf{0.2481 \pm 0.0007}$ & $0.2655 \pm 0.0003$ & $0.2643 \pm 0.0008$\\
\hline
Computer & 	$90.34 \pm 0.45$ & $91.48 \pm 0.41$ & $\bf{91.59 \pm 0.31} $ & $91.43 \pm 0.53$\\
\bottomrule
\end{tabular}
}
\end{table*}

\subsection{Long-Range Graph Benchmark}
We have additionally conducted a set of experiments on the Long-Range Graph Benchmark (LRGB, \citealp{lrgb-2022}), which consists of five challenging datasets that test a model's ability to learn \emph{long-range dependencies} in input graphs. The results are presented in \Cref{tab:lrgb}, which shows that our \emph{\textsc{Exphormer}-based sparse attention models are able to outperform GraphGPS on three of the five datasets, achieving SOTA performance}. Furthermore, on the remaining two datasets in the LRGB suite, \textsc{Exphormer} is competitive with the best single model results, obtained by full attention GraphGPS models.

\subsection{Scaling to Larger Graphs}
One of the shortcomings of graph transformer architectures has been their poor scalability to larger graphs on thousands of nodes. Dense attention mechanisms (e.g., SAN and Graphormer) have quadratic memory complexity and time complexity, which restrict their applicability to datasets on graphs with relatively few nodes, such as molecular graphs.

GraphGPS has made use of sparse attention mechanisms, but even so, the architecture handles graphs of up to about 5,000 nodes (e.g., MalNet-Tiny,~\citealp{FreitasDNC21}), often with very small batch size (see \Cref{rmk:batchsize}).


A natural question is whether the sparse attention mechanism of \textsc{Exphormer} models allows us to extend graph transformer architectures to significantly larger graphs while providing competitive performance. Indeed, we show this is the case by training \textsc{Exphormer} models on larger datasets, including Amazon Computer, Amazon Photo, Coauthor CS, Coauthor Physics~\citep{amazon_dataset}, which has up to 35K nodes and 250K edges. We also show the use of \textsc{Exphormer} on ogbn-arxiv, which has 169K nodes and 1.1M edges. The results are shown in \Cref{tab:largegraphs}. Standard GraphGPS models suffer from Out Of Memory (OOM) issues on a number of these datasets, demonstrating the utility of \textsc{Exphormer}. To provide meaningful comparisons to other transformer architectures, we thus include NAGphormer~\citep{nagphormer22}, a scalable sampling-based graph transformer architecture, as a baseline. \textsc{Exphormer} performs competitively and, in fact, achieves SOTA accuracy on Computer: the best non-transformer method has an accuracy of 90.74~\citep{computersSOTA}.

We would also like to highlight that \textsc{Exphormer} can be used for even larger graphs, with hundreds of thousands of nodes. In particular, we provide results on ogbn-arxiv~\citep{HuFZDRLCL20}, a challenging transductive dataset consisting of a single large graph of the citation network of over 160K arXiv papers, containing over a million edges.
Specifically, we achieve a test accuracy of {\bf 0.7244} using the \textsc{Exphormer} architectures. At the time of writing, a relevant leaderboard~\citep{ogbleaderboard} shows 0.7966 as the highest reported test accuracy~\citep{zhao:variational}. \Cref{tab:arxiv} compares \textsc{Exphormer} to other MPNNs and sparse transformers. 
A dense full transformer does not even fit in memory on a 40GB NVIDIA A100 GPU (even for a tiny network of only 2 layers and 32 hidden dimensions). We then fixed the network size to 3 hidden layers and 96 hidden dimensions to be able to fit the sparse models in memory. Notice that BigBird and Performer have significantly longer epoch times and worse performance than with just a MPNN. \textsc{Exphormer} with virtual nodes only (without expander edges) improves on the MPNN and provides an accuracy of 0.7222. \textsc{Exphormer} with expander edges further improves the accuracy to 0.7244.

\subsection{Ablation Studies}
\label{sec:ablation}
Here we analyze the effect of each of the components of the model. Our \textsc{Exphormer} model has three main components: local neighborhood, expander edges, and virtual nodes. In \cref{tab:alations_comps}, we analyze which parts of the model have been useful, and which parts can be removed without reducing the performance. In all of these experiments, the MPNN part is fixed, so it gives a baseline number for the results, and the transformer part boosts over the MPNN part.
Through these experiments we can see that local neighborhood have been always a good option to add to the \emph{Exphormer}, but between virtual nodes and expander graphs, sometimes one of them causes reduction in the performance.
This issue is discussed further in \cref{sec:expander-vs-global}.

\section{Conclusion}

\textsc{Exphormer} is a new class of sparse graph transformer architectures. We introduced two types of sparse networks with virtual nodes and using expander graphs. We have shown that the mathematical properties of our architecture make \textsc{Exphormer} a suitable choice for graph learning. Our sparse architecture uses fewer training parameters, is faster to train and has memory complexity linear in the size of the graph.  These properties help us scale to larger graphs which were typically elusive to other Transformer-based methods. \textsc{Exphormer} outperforms other sparse transformers and performs comparably or better than full transformers. We have shown the applicability of \textsc{Exphormer} on a wide variety of graph learning datasets. Combining \textsc{Exphormer} with MPNN in the GraphGPS framework allows us to obtain state-of-the-art empirical results on a number of datasets. 

\section*{Acknowledgements}
This work was supported in part by the Natural Sciences and Engineering Resource Council of Canada, the Canada CIFAR AI Chairs program, the BC DRI Group, Compute Ontario, Calcul Qu\'ebec, and the Digital Resource Alliance of Canada.

\bibliography{bibliography}
\bibliographystyle{icml2023}

\newpage
\appendix
\onecolumn

\section{Dataset Descriptions} \label{sec:datasetdesc}
Below, we provide descriptions of the datasets on which we conduct experiments.

\paragraph{CIFAR10 and MNIST} \citep{DwivediJLBB20}
CIFAR10 and MNIST are the graph equivalents of the image classification datasets of the same name. A graph is created by constructing the 8-nearest neighbor graph of the SLIC superpixels of the image. These are both 10-class graph classification problems.

\paragraph{PascalVOC-SP and COCO-SP} \citep{dwivedi2021graph}
These are similar graph versions of image datasets, but they are larger images and the task is to perform node classification, i.e.\ semantic segmentation of superpixels.
These graphs are larger, and the task more complex, than CIFAR10 and MNIST.

\paragraph{CLUSTER and PATTERN} \citep{DwivediJLBB20}
PATTERN and CLUSTER are node classification problems. Both are synthetic datasets that are sampled from a Stochastic Block Model (SBM), is a popular way to model communities. In PATTERN, the prediction task is to identify if a node belongs to one of the 100 possible predetermined subgraph patterns.  In CLUSTER, the goal is to classify nodes into six different clusters with the same distribution.

\paragraph{MalNet-Tiny} \citep{FreitasDNC21}
Malnet-Tiny is a smaller dataset generated from a larger dataset for identifying malware based on function call graphs from Android APKs. The tiny dataset contains 5000 graphs, each with up to 5000 nodes. The task is to predict the graph as being benign or from one of four types of malware.



\paragraph{ogbn-arxiv} \citep{ogbPaper}
The ogbn-arxiv dataset consists of one large
directed graph of 169343 nodes and 1,166,243 edges representing a citation network between all
computer science papers on arXiv that were indexed by the
Microsoft academic graph. Nodes in the graph represent
papers, while a directed edge indicates that a paper cites
another. Each node has an 128-dimensional feature vector
derived from embeddings of words in the title and abstract
of the underlying paper. The prediction task is a 40-class
node classification problem --- to identify the primary
category of each arXiv paper, as listed by the authors.
Moreover, the nodes of the citation graph are split into
90K training nodes, 30K validation notes, and  48K test nodes.

\paragraph{Coauthor datasets} 
Coauthor CS and Physics are co-authorship graphs from Microsoft Academic Graph.  The nodes represent the authors
and two authors who share a paper are connected by an edge.  The node features are from the keywords in the papers.
The class represent the active area of study for the author.

\paragraph{Amazon datasets}
Amazon Computers and Amazon photo are Amazon co-purchase graphs. Nodes represents products purchased an edges indicate
pairs of products purchased together.  Node features are bag-of-words encoded reiews of the products. Class labels are the product category.

\paragraph{Peptides-Func, Peptides-Struct, and PCQM-Contact} \citep{dwivedi2021graph} These datasets are molecular graphs introduced as a part of the Long Range Graph Benchmark (LRGB). Graphs in these datasets have relatively large diameters: the average diameter for PCQM-Contact is $9.86 \pm 1.79$, and for the Peptides datasets $56.99 \pm 28.72$. Average shortest path lengths are $4.63 \pm 0.63$ and  $20.89 \pm 9.79$ accordingly. On PCQM-Contact the task is edge-level, and we need to rank the edges. Peptides-Func is a multi-label graph classification task, with 10 labels. Peptides-Struct is graph-level regression of 11 structural properties of the molecules.

\Cref{table:datasets} shows a summary of the statistics of the
aforementioned datasets.

\begin{table}[htp]
    \centering
    \caption{Dataset statistics}
    \fontsize{8.5pt}{8.5pt}\selectfont
    \setlength\tabcolsep{6pt} 
    \scalebox{1}{
    \begin{tabular}{lcccccc}
    \toprule
         {\bf Dataset} & {\bf Graphs} & {\bf Avg. nodes} & {\bf Avg. edges} & {\bf Prediction Level} & {\bf No. Classes} &{\bf Metric} \\
    \midrule
MNIST & 70,000&  70.6 & 564.5 & graph& 10 & Accuracy\\
CIFAR10 & 60,000 & 117.6 & 941.1 &  graph & 10 & Accuracy\\
PATTERN & 14,000 & 118.9 & 3,039.3  & inductive node & 2 & Accuracy\\
CLUSTER & 12,000 & 117.2 & 2,150.9  & inductive node & 6 & Accuracy\\
MalNet-Tiny & 5,000 & 1,410.3 & 2,859.9 & graph & 5 & Accuracy\\
\midrule
PascalVOC-SP & 11,355 & 479.4 & 2,710.5 & inductive node & 21 & F1\\
COCO-SP & 123,286 & 476.9 & 2,693.7 & inductive node & 81 & F1\\
PCQM-Contact & 529,434 & 30.1 & 61.0 & inductive link & (link ranking) & MRR\\
Peptides-func & 15,535 & 150.9 & 307.3 & graph & 10 & Average Precision\\
Peptides-struct & 15,535 & 150.9 & 307.3 & graph & 11 (regression) & Mean Absolute Error\\
\midrule
ogbn-arxiv & 1 & 169,343 & 1,166,243 & node & 40 & Accuracy\\
Amazon Computer & 1  & 13381 & 245778 & node & 10 & Accuracy \\
Amazon Photo &2 & 7487 & 119043 & node & 8 & Accuracy \\
Coauthor CS & 1 & 18333 & 81894 & node & 15 & Accuracy \\
Coauthor Physics & 1 & 34493 & 247962 & node & 5 & Accuracy \\
    \bottomrule
    \end{tabular}
    }
     \label{table:datasets}
\end{table}

\section{More Experimental Results}

\subsection{Hyperparameters}
Our hyperparameter choices, including the optimizer, positional encodings, and structural encodings, were guided by the instructions in GraphGPS \citep{RampasekGDLWB22}. There were some cases, however, when more layers with smaller dimensions gave better results in \textsc{Exphormer}. This may be due to the fact that each node gets fewer inputs for each layer, but \textsc{Exphorme} requires more layers in order to propagate well. Additionally, we observed that Equivstable Laplacian Positional Encoding (ESLapPE) performed better than normal Laplacian Positional Encoding (LapPE) in some cases, in cases we replaced you can see ESLapPE version of GraphGPS results in \Cref{tab:ablationcifar,tab:MNIST,tab:cluster,tab:pattern}. Except for the large scale graphs, which we use GCN, we have used CustomGatedGCN beside the Exphormer in all of the experiments.

Through our model, some extra hyperparameters are introduced --- the degree of the graph expander and the number of virtual nodes. For these hyperparameters, we used linear search and found that expander degree 6-22 was the most effective. Depending on the graph size, we used 1-6 virtual nodes. As the number of hyperparameters is large, grid search was not feasible on all the parameters. As a result, for other hyperparameters, we did a linear search for each parameter to find out which parameters work better.

To make fair comparisons, we used a similar parameter-budget to GraphGPS. For PATTERN and CLUSTER, we used a parameter-budget of 500K, and for CIFAR and MNIST, we used a parameter-budget of around 100K. See details in \Cref{tab:hyperparams-gps-1,tab:hyperparams-lrgb,tab:hyperparams-large}.

\begin{table}[ht]
\centering
\caption{Hyperparameters used for \textsc{Exphormer} for datasets: CIFAR10, MNIST, MalNet-Tiny, PATTERN, CLUSTER.}
\label{tab:hyperparams-gps-1}
\begin{tabular}{l|lllll} 
\toprule
{\bf Hyperparameter}    & {\bf CIFAR10} & {\bf MNIST}   & {\bf MalNet-Tiny} & {\bf PATTERN} & {\bf CLUSTER}  \\ 
\hline
Num Layers        & 5       & 5       & 5           & 4       & 20       \\
Hidden Dim        & 40      & 40      & 64          & 40      & 32       \\
Num Heads         & 4       & 4       & 4           & 4       & 8        \\
Dropout           & 0.1     & 0.1     & 0           & 0.0     & 0.0      \\
PE                & ESLapPE & ESLapPE & None        & ESLapPE & ESLapPE  \\ 
\midrule
Batch Size        & 16      & 16      & 16          & 32      & 16       \\
Learning Rate     & 0.001   & 0.001   & 0.0005      & 0.0002   & 0.0002  \\
Num Epochs        & 150     & 150     & 150         & 120     & 200      \\ 
\midrule
Expander Degree   & 10       & 10       & -           & 14       & 6        \\
Num Virtual Nodes & 1       & 1       & 4           & 4       & 3        \\
Num parameters    & 111,095 & 111,015 & 286,277     & 91,045 & 282,970  \\
\bottomrule
\end{tabular}

\end{table}

\begin{table}[ht]
\centering
\caption{Hyperparameters used for \textsc{Exphormer} for LRGB datasets.}
\label{tab:hyperparams-lrgb}
\begin{tabular}{l|lllll} 
\toprule
{\bf Hyperparameter}    & {\bf PascalVOC-SP} & {\bf COCO-SP}   & {\bf Peptides-Func} & {\bf Peptides-Struct} & {\bf PCQM-Contact}  \\ \hline
Num Layers        & 4       & 7       & 8           & 4       & 7       \\
Hidden Dim        & 96      & 72      & 64          & 88      & 64       \\
Num Heads         & 8       & 4       & 4           & 4       & 4        \\
Dropout           & 0.15    & 0       & 0.12        & 0.12    & 0      \\
PE                & LapPE   & LapPE   & LapPE       & LapPE   & LapPE  \\ 
\hline
Batch Size        & 32      & 32      & 128         & 128     & 128       \\
Learning Rate     & 0.0005  & 0.0005  & 0.0003      & 0.0003  & 0.0003    \\
Num Epochs        & 300     & 300     & 200         & 200     & 200      \\ 
\hline
Expander Degree   & 14       & 22       & -           & -       & -        \\
Num Virtual Nodes & 0       & 0       & 1           & 6       & 6        \\
Num parameters    & 509,301 & 498,993 & 445,866     & 426,427 & 395,936  \\
\bottomrule
\end{tabular}

\end{table}

\begin{table}[ht]
\centering
\caption{Hyperparameters used for \textsc{Exphormer} for large transductive graph datasets.}
\label{tab:hyperparams-large}
\begin{tabular}{l|lllll} 
\toprule
{\bf Hyperparameter}    & {\bf OGBN-Arxiv} & {\bf Computer}   & {\bf Photo} & {\bf CS} & {\bf Physics}  \\ \hline
Num Layers              & 3       & 4       & 4           & 4       & 4       \\
Hidden Dim              & 96      & 80      & 64         & 72      & 72       \\
Num Heads               & 2       & 2       & 2           & 2       & 2        \\
Dropout                 & 0.3     & 0.4    & 0.4         & 0.4     & 0.4      \\
PE                      & -       & -       & -           & -   & -  \\ 
\hline
Learning Rate           & 0.01    & 0.001   & 0.001       & 0.001  & 0.001    \\
Num Epochs              & 600     & 150     & 100         & 70     & 70      \\ 
\hline
Expander Degree         & 6       & 6       & 6           & 6       & 6        \\
Num Virtual Nodes       & 0       & 0       & 0           & 0       & 0        \\
Num parameters          & 268,264 & 302,570 & 202,632     & 686,103 & 801,293  \\
\bottomrule
\end{tabular}

\end{table}

\subsection{Full Comparison of Attention Mechanisms} \label{app:attncomp}
\begin{remark}
\textsc{Exphormer} has some conceptual similarities with BigBird, as mentioned previously. For instance, we also make use of virtual global attention nodes, corresponding to \textsc{BigBird-etc}.

However, our approach departs from that of BigBird in some important ways. While BigBird uses $w$-width ``window attention'' to capture locality of reference, we use local neighborhood attention to capture locality and graph topology. In particular, the interaction graph due to window attention in BigBird can be viewed as a Cayley graph on $\Z_n$, which is sequence-centric, while \textsc{Exphormer} is graph-centric and, therefore, uses the structure of the input graph itself to capture locality.
BigBird, as implemented for graphs by \citet{RampasekGDLWB22}, instead simply orders the graph nodes in an arbitrary sequence and uses windows within that sequence.

Both BigBird and \textsc{Exphormer} also make use of a random attention model. While BigBird uses an Erd\H{o}s-R\'{e}nyi graph on $|V|$ nodes, our approach is to use a $d$-regular expander for fixed constant $d$. The astute reader may recall that a Erd\H{o}s-R\'{e}nyi graph $G(n,p)$ has spectral expansion properties for large enough $p$. However, it is known that $p = \frac{\log n}{n}$ is the connectivity threshold, i.e., for $p < (1-\epsilon) \frac{\log n}{n}$, $G(n,p)$ is almost surely a disconnected graph. Therefore, in order to obtain even a connected graph in the Erd\H{o}s-R\'{e}nyi model -- let alone one with expansion properties -- one would need $p = \Omega\left(\frac{\log n}{n}\right)$, giving superlinear complexity for the number of edges.
BigBird uses $p = \Theta(1/n)$, keeping a linear number of edges but losing expansion properties.
Our expander graphs, by contrast, allow both a linear number of edges and guaranteed spectral expansion properties.

We will see in the practical experiments of \cref{sec:experiments} that \textsc{Exphormer}-based models often substantially outperform BigBird-based equivalents, with fewer parameters.
\end{remark}

In \Cref{sec:attentioncomp}, we presented two approaches for the comparison of models trained using different attention mechanisms --- fixing the hyperparameters and fixing a budget on the total number of trainable parameters. The results showed the advantage of \textsc{Exphormer} over other attention mechanisms for CIFAR10 (\Cref{tab:ablationcifar}) and PATTERN (\Cref{tab:pattern}). Here, we present similar results for the remaining datasets --- MNIST in \cref{tab:MNIST}; MalNet-Tiny in \Cref{tab:malnet}; PATTERN in \Cref{tab:pattern}; and CLUSTER in \Cref{tab:cluster}.

\begin{table*}[t!]
    \centering
    \caption{Results with varying attention and MPNNs on CIFAR10. \textsc{Exphormer} with MPNN provides the highest accuracy. Also, pure transformer models based on \textsc{Exphormer} (without the use of MPNNs) are comparable.}
\label{tab:ablationcifar}
    \setlength\tabcolsep{7pt} 
    \scalebox{0.8}{
    \begin{tabular}{lcccccca}
    \toprule
         {\bf Model} & {\bf \#Layers} & {\bf Hidden} & {\bf \#Positional} & {\bf Expander} & {\bf \#Parameters} & {\bf Time}& {\bf Accuracy} \\
          & & {\bf layers} & {\bf encoding} & {\bf degree} & & {\bf Epch/Total} & {\bf } \\
     \midrule
         GPS-MPNN: GatedGCN & 3 & 52 & LapPE & - & 79,654 & 43s/1.18h & $ 69.95\pm0.499$ \\
         GPS: Transformer & 3 & 52 & LapPE & - & 70,762 & 40s/1.11h & $68.86\pm1.138$\\
         GPS: Transformer + MPNN & 5 & 40 & ESLapPE & - & 111,735 & 104s/2.89h & $73.53 \pm 0.238$ \\
         GPS: Transformer + MPNN & 3 & 52 & LapPE & - & 112,726 & 62s/1.72h & $72.31 \pm 0.344$ \\
         GPS: Performer + MPNN & 5 & 40 & ESLapPE & - & 283,935 & 132s/3.65h & $70.18 \pm0.095$ \\
         GPS: Performer + MPNN & 3 & 52 & LapPE & - & 239,554 & 77s/2.14h & $70.67 \pm 0.338$ \\
         GPS: BigBird + MPNN & 5 & 40 & ESLapPE & - & 128,335 & 243s/6.75h & $70.51 \pm 0.256$ \\
         GPS: BigBird + MPNN & 3 & 52 & LapPE & - & 129,418 & 145s/4h & $70.48 \pm 0.106$ \\
     \midrule
         \textsc{Exphormer} w/o MPNN& 5 & 44 & ESLapPE & 10 & 84,134 & 64s/1.78h & $72.33 \pm0.155$\\
         \textsc{Exphormer} w/o MPNN& 7 & 44 & ESLapPE & 10 & 119,022 & 80s/2.23h & $72.88 \pm0.166$ \\
         \textsc{Exphormer} & 5 & 40 & ESLapPE & 10 & 111,095 & 115s/3.21h & $74.75 \pm 0.194$ \\
         \textsc{Exphormer} & 5 & 44 & ESLapPE & 10 & 133,819 & 114s/3.19h& \bf{75.03 $\pm$ 0.186} \\
     \bottomrule
    \end{tabular}
    }

\end{table*}

\begin{table}[ht]
    \centering
    \caption{Ablation studies results for MNIST}
    \label{tab:MNIST}
    \scalebox{0.8}{
    \begin{tabular}{lcccccca}
    \toprule
         {\bf Model} & {\bf \#Layers} & {\bf Hidden} & {\bf \#Positional} & {\bf Expander} & {\bf \#Parameters} & {\bf Time} & {\bf Accuracy} \\
          & & {\bf layers} & {\bf encoding} & {\bf degree} & & {\bf Epoch/Total} & {\bf } \\
     \midrule
         GPS: Transformer + MPNN & 5 & 40 & ESLapPE & - & 111,655 & 131s/5.45h & $98.336 \pm 0.0189$ \\
         GPS: Transformer + MPNN & 3 & 52 & LapPE & - & 115,394 & 76s/2.13h & $98.051 \pm 0.126$ \\
         GPS: Performer + MPNN & 5 & 40 & ESLapPE & - & 283,855 & 156s/6.52h & $98.34 \pm 0.0349$ \\
         GPS: BigBird + MPNN & 5 & 40 & ESLapPE & - & 128,255 & 267s/11.11h & $98.176\pm0.0146$ \\
     \midrule
         \textsc{Exphormer} w/o MPNN & 5 & 44 & ESLapPE & 10 & 92,146 & 75s/3.14h & $98.08\pm0.051$ \\
         \textsc{Exphormer} w/o MPNN & 7 & 44 & ESLapPE & 10 & 127,698 & 93s/3.87h & $98.238\pm0.0387$  \\
         \textsc{Exphormer} & 5 & 40 & ESLapPE & 10 & 111,015 & 132s/5.49h & $98.414 \pm 0.035$ \\
         \textsc{Exphormer} & 5 & 44 & ESLapPE & 10 & 133,731 & 137s/5.72h & \bf{98.424 $\pm$ 0.018} \\
     \bottomrule
    \end{tabular}
    }
\end{table}

\begin{table}[ht]
    \centering
    \caption{Ablation studies results for MalNet-Tiny. We want to remark that these results are based on one virtual node. For the best result reported in the main paper, we have used 4 virtual nodes, which led to much better numbers. The model marked with * did not fit in memory with batch size 16, and was trained with batch size 8.}
    \label{tab:malnet}
    \label{tab:malnettiny}
    \scalebox{0.8}{
    \begin{tabular}{lcccccca}
    \toprule
         {\bf Model} & {\bf \#Layers} & {\bf Hidden} & {\bf \#Positional} & {\bf Expander} & {\bf \#Parameters} & {\bf Time} & {\bf Accuracy} \\
          & & {\bf layers} & {\bf encoding} & {\bf degree} & & {\bf Epoch/Total} & {\bf } \\
     \midrule
         GPS-MPNN: GatedGCN & 5 & 64 & - & - & 199,237 & 6s/0.25h & $92.23\pm0.65$ \\
         GPS: Performer & 5 & 64 & - & - & 421,957 & 41s/1.73h & $73.90\pm0.58$ \\
         GPS: Transformer + MPNN* & 5 & 64 & - & - & 282,437 & 94s/3.94h & \bf{93.50 $\pm$ 0.41} \\
         GPS: Performer + MPNN & 5 & 64 & - & - & 527,237 & 46s/1.90h & $92.64 \pm 0.78$ \\
         GPS: BigBird + MPNN & 5 & 64 & - & - & 324,357 & 130s/5.43h & $92.34 \pm 0.34$ \\
     \midrule
         \textsc{Exphormer} w/o MPNN & 5 & 80 & - & 10 & 283,173 & 25.2s/1.05h & $92.18 \pm 0.292$ \\
         \textsc{Exphormer} w/o MPNN & 8 & 64 & - & 10 & 296,325 & 35.2s/1.47h & $92.24 \pm0.291$ \\
         \textsc{Exphormer} & 5 & 64 & - & 10 & 286,277 & 25.1s/1.05h & $94.02 \pm 0.209$ \\
     \bottomrule
    \end{tabular}
    }
\end{table}

\begin{table}[htp]
    \centering
    \caption{Ablation studies results for PATTERN}
    \label{tab:pattern}
    \scalebox{0.8}{
    \begin{tabular}{lcccccca}
    \toprule
         {\bf Model} & {\bf \#Layers} & {\bf Hidden} & {\bf \#Positional} & {\bf Expander} & {\bf \#Parameters} & {\bf Time} & {\bf Accuracy} \\
          & & {\bf dimension} & {\bf encoding} & {\bf degree} & & {\bf Epoch/Total} & {\bf } \\
     \midrule
         GPS: Transformer + MPNN & 4 & 40 & ESLapPE & - & 91,165 & 18s/0.59h & $86.114 \pm 0.103$  \\
         GPS: Transformer + MPNN & 6 & 64 & LapPE-16 & - & 337,201 & 32s/0.89h & $86.685 \pm 0.059$ \\
         GPS: Performer + MPNN & 4 & 40 & ESLapPE & - & 228,925 & 20s/0.68h & $83.342 \pm 0.294$ \\
         GPS: BigBird + MPNN & 4 & 40 & ESLapPE & - & 104,445 & 29s/0.97h & $86.005 \pm 0.148$ \\
     \midrule
         \textsc{Exphormer} w/o MPNN & 4 & 40 & ESLapPE & 14 & 56,801 & 13s/0.44h & $85.622 \pm 0.007$\\
         \textsc{Exphormer} w/o MPNN & 4 & 56 & ESLapPE & 14 & 109,985 & 14s/0.47h & $85.601 \pm 0.009$ \\
         \textsc{Exphormer} & 4 & 40 & ESLapPE & 14 & 91,045 & 21s/0.71h & \bf{86.734 $\pm$ 0.008} \\
     \bottomrule
    \end{tabular}
    }
\end{table}

\begin{table}[hb]
    \centering
    \caption{Ablation studies results for CLUSTER}
    \label{tab:cluster}
    \scalebox{0.8}{
    \begin{tabular}{lcccccca}
    \toprule
         {\bf Model} & {\bf \#Layers} & {\bf Hidden} & {\bf \#Positional} & {\bf Expander} & {\bf \#Parameters} & {\bf Time} & {\bf Accuracy} \\
          & & {\bf dimension} & {\bf encoding} & {\bf degree} & & {\bf Epoch/Total} & {\bf } \\
     \midrule
         GPS: Transformer + MPNN & 20 & 32 & ESLapPE & - & 285,210 & 91s/3.79h & $78.053 \pm 0.044$ \\
         GPS: Transformer + MPNN & 16 & 48 & LapPE-10 & - & 502,054 & 86s/2.40h & $78.016 \pm 0.180$ \\
         GPS: Performer + MPNN & 20 & 32 & ESLapPE & - & 1,512,090 & 120s/5.01h & \bf{78.292 $\pm$ 0.046} \\
         GPS: BigBird + MPNN & 20 & 32 & ESLapPE & - & 328,090 & 224s/9.32h & $77.468 \pm 0.023$ \\
     \midrule
         \textsc{Exphormer} w/o MPNN & 20 & 32 & ESLapPE & 6 & 171,590 & 55s/3.06h & $76.922 \pm 0.044$\\
         \textsc{Exphormer} & 20 & 32 & ESLapPE & 6 & 282,970 & 102s/5.69h & $78.07 \pm 0.037$ \\
     \bottomrule
    \end{tabular}
    }
    
\end{table}

\begin{table}[htbp]
    \centering
    \caption{Comparison of attention mechanisms on the ogbn-arxiv dataset by fixing the network size.}
    \label{tab:arxiv}
    \fontsize{8.5pt}{8.5pt}\selectfont
    \scalebox{1}{
    \begin{tabular}{lccc}
    \toprule
         {\bf Model} & {\bf Accuracy} & {\bf Epoch times} & {\bf \#Parameters} \\
         \midrule
         \rowcolor{lightyellow} \textsc{GCN+Exphormer} with expander edges &  72.44 $\pm$ 0.28 & 1.72 & 268,552  \\
         \rowcolor{lightyellow} \textsc{GCN+Exphormer} with virtual nodes & 72.22 $\pm$ 0.13 & 1.76 & 269,704 \\
         GCN only & 71.35 $\pm$ 0.31 & 0.21 &  74,152 \\
         GCN+BigBird & 71.16 $\pm$ 0.19 & 17.85 &  325,480 \\
         GCN+Performer &  70.92 $\pm$ 0.04 & 5.77 & 452,776 \\
 \bottomrule
    \end{tabular}
}
    \vspace{-0.1in}
\end{table}

\section{Details of Expander Graph Construction} \label{sec:expander-details}
A major component of \textsc{Exphormer} is the use of the edges of an expander graph. In this section, we provide details of the specific expander graphs we use as well and quantify their spectral expansion  properties.

\subsection{Ramanujan Graphs}
A natural question is how strong the spectral expansion properties of a $d$-regular graph can be, i.e., for how large an $\epsilon > 0$ does a $d$-regular $\epsilon$-expander exist. The following theorem gives a bound on how large the spectral gap can be.
\begin{theorem}[Alon-Boppana]
 Let $d > 0$. The eigenvalues of the adjacency matrix of a $d$-regular graph on $n$ nodes satisfy 
 \[
   \max\{|\lambda_2|, |\lambda_n|\} \geq 2\sqrt{d-1} - o_n(1).
 \]
 In other words, a $d$-regular $\epsilon$-expander graph can exist only for $\epsilon \geq \frac{2\sqrt{d-1}}{d} - o_n(1)$.
\end{theorem}

As it turns out, there exist $\epsilon$-expander graphs with $\epsilon$ achieving this bound. In fact, a $d$-regular $\epsilon$-expander graph satisfying $\epsilon \leq \frac{2\sqrt{d-1}}{d}$ is known as a \emph{Ramanujan graph}. Ramanujan graphs are essentially the best possible spectral expanders, and several constructions have been considered over the years~\citep{LubotzkyPS88, Margulis88}.

\subsection{Random Regular Graphs}
While there exist deterministic constructions of Ramanujan graphs, they are often algebraic/number theoretic in nature and therefore exist only for specific choices of $d$ (e.g., the constructions of \citet{LubotzkyPS88} as well as independently of \citet{Margulis88}, for which one requires $d\equiv 2\pmod{4}$ and $d-1$ to be a prime). Recently, the work of \citet{Alon21} showed a construction of strongly explicit near-Ramanujan graphs of every degree, but it should be noted that the construction needs the number of nodes to be sufficiently large. It is, therefore, often convenient to use a probabilistic construction of an expander.

Below, we consider three different probabilistic constructions of expander graphs, which are compared in \Cref{tab:expander_algorithms}.

\paragraph{Standard expander graph construction}
A natural choice for an expander graph is a \emph{random} $d$-regular graph on $n$ vertices, formed by taking $d/2$ independent uniform permutations on $\{1, 2, \dots, n\}$. \citet{Friedman03} proved a conjecture of Alon, establishing that random regular graphs are \emph{weakly-Ramanujan}.
\begin{theorem}\citep[Theorem 1.1]{Friedman03}
\label{thm:randomregular}
 Fix $\epsilon > 0$ and an even integer $d > 2$. Then, suppose $G = (V,E)$ is a random graph generated by taking $d/2$ independent uniformly random permutations $\pi_1, \pi_2, \dots, \pi_{d/2}$ on $V = \{1,2,\dots, n\}$ and then choosing the edge set as
 \[
   E = \left\{ (i, \pi_j(i)), (i, \pi_j^{-1}(i)): 1\leq j\leq d, 1\leq i\leq n\right\}.
 \]
 Then, with probability $1 - O(n^{-\Omega(\sqrt{d})})$,
 $G$ satisfies
 $\lambda_j(G) \leq 2\sqrt{d-1}+\epsilon$ for $j=2,\dots, n$, where $d = \lambda_1(G) \geq \lambda_2(G) \geq \cdots \geq \lambda_n(G) \geq -d$ are the eigenvalues of the adjacency matrix of $G$. 
\end{theorem}

In the experiments reported in \cref{tab:benchmarking,tab:sparse-comparision,tab:lrgb,tab:largegraphs,tab:alations_comps}, we use graphs generated according to the random process described above, with self-loops removed, to instantiate the expander graph component of $\textsc{Exphormer}$.

\paragraph{A simple variant.} A simple variant of the aforementioned random process is to instead choose a single permutation on $nd/2$ elements formed by taking $d/2$ copies of $1,2,\dots, n$. More precisely, given an input graph $(V,E)$, one can generate an expander graph $(V, E')$ as follows:
\begin{itemize}
    \item Let $s = (\underbrace{1, \dots, 1}_{d/2}, \underbrace{2, \dots, 2}_{d/2}, \dots, \underbrace{n, \dots, n}_{d/2})$.
    \item Pick a permutation $\pi$ on $\{1,2,\dots, n d/2\}$ uniformly at random from the $(nd/2)!$ possible permutations.
    \item Let $E'$ be the multiset $\{ (s_i, s_{\pi(i)}), (s_{\pi(i)}, s_i):  1\leq i\leq nd/2\}$.
\end{itemize}
For the ablation studies in \cref{tab:ablationcifar,tab:MNIST,tab:malnet,tab:pattern,tab:cluster}, we use the above variant to generate the expander graph component of $\textsc{Exphormer}$ (once again removing self loops). We find that, in practice, the above variant produces near-Ramanujan graphs most of the time.

\paragraph{Hamiltonian cycle variant.} Another interesting variant of the random graph process analyzed in \cref{thm:randomregular} is the one in which one once again takes $d/2$ independent random permutations, except that the permutations must be one of the $(n-1)!$ permutations consisting of a single cycle of length $n$. In other words, one chooses $d/2$ independent Hamiltonian cycles on $V$ and takes all edges from each of the Hamiltonian cycles. The work of \citet{Friedman03}, in fact, analyzes this modified process:

\begin{theorem}\citep[Theorem 1.2]{Friedman03}
 Fix $\epsilon > 0$ and an even integer $d > 2$. Then, suppose $G = (V,E)$ is a random graph generated as follows: Take independent permutations $\pi_1, \pi_2, \dots, \pi_{d/2}$ on $V = \{1,2,\dots, n\}$, where each $\pi_i$ is chosen uniformly at random from the $(n-1)!$ permutations whose cyclic decomposition consists of a single cycle. Then choose the edge set as
 \[
   E = \left\{ (i, \pi_j(i)), (i, \pi_j^{-1}(i)): 1\leq j\leq d, 1\leq i\leq n\right\}.
 \]
 Then, with probability $1 - O(n^{-\Omega(\sqrt{d})})$,
 $G$ satisfies
 $\lambda_j(G) \leq 2\sqrt{d-1}+\epsilon$ for $j=2,\dots, n$, where $d = \lambda_1(G) \geq \lambda_2(G) \geq \cdots \geq \lambda_n(G) \geq -d$ are the eigenvalues of the adjacency matrix of $G$. 
\end{theorem}
The advantage of using this random process for generating an expander graph is that it automatically satisfies the universal approximation property outlined in \cref{sec:proofs} when augmented with self-loops (see \cref{thm:univapprox}).

\paragraph{Experimental analysis:} In \cref{tab:expander_algorithms}, we have compared the effect of using different algorithms for expander graph creation. We can see from the results that all approaches have almost similar results. Note that \cref{thm:univapprox} assumes the presence of Hamiltonian paths for its approximation properties.

\begin{table}
\centering
\refstepcounter{table}
\label{tab:expander_algorithms}
\caption{An experimental comparison of different approaches used for Ramanujan Expander graph generation.}
\scalebox{0.8}{
\begin{tabular}{l|lllll} 
\toprule

Expander Algorithm & Cifar10       & MNIST         & Cluster       & Pattern       & PascalVOC-SP     \\ 

\hline

Standard Permutation-Based               & $74.69 \pm 0.125$ & $98.55 \pm 0.039$ & $78.07 \pm 0.037$ & $86.74 \pm 0.015$ & $0.3975 \pm 0.0037$  \\

Simple Permutation-Based Variant        & $74.75 \pm 0.194$ & $98.41 \pm 0.038$ & $78.12 \pm 0.030  $ & $86.73 \pm 0.008$ & $0.3966 \pm 0.0027$  \\

Hamiltonian Cycle Variant         & $74.73 \pm 0.099$ & $98.40 \pm 0.032$ & $78.01 \pm 0.027$ & $86.73 \pm 0.012$ & $0.3990 \pm 0.0037$  \\

\bottomrule
\end{tabular}
}
\end{table}

\section{On Expander Graphs versus Global Connectors} \label{sec:expander-vs-global}
From ablation studies in \Cref{sec:ablation}, we can see that sometimes only having expander graphs leads to the best results, sometimes virtual nodes only, and sometimes using a combination of both. Although this is a hyperparameter to tune, we can have a general understanding from the datasets on which cases virtual nodes work better and vice versa. Here we remark on some of these patterns.

\textbf{Graph diameter}: Using virtual nodes, the diameter of the graph becomes exactly 2 (unless the original graph was already complete, with diameter 1). Expander graphs also help to reduce the diameter of the graph, but only achieve a (probabilistic) diameter guarantee of $O(\log n)$.

\textbf{Information Bottleneck}: Virtual nodes serve as a global sink, and so even a single node should be able to keep all the information flowing from the whole graph. In case the graphs are large or many nodes rely on the virtual nodes to pass information, though, virtual nodes are likely to cause information bottleneck, and may even reduce the performance of the model. Expander graphs, to the contrary, introduce many new paths; as they rely on the local information storage of the nodes, they do not have this same problem.

\textbf{Interference with the local structure}: Virtual nodes introduce a new node, through which all of their new connections pass. Expander edges, on the other hand, are much easier to confuse with the actual edges of the graph; this can make it more difficult for the model to use the graph structure appropriately. As our model uses shared weights among the types of activations, which substantially helps with model size, the only way to understand the edges are different is through the edge embeddings. For datasets like MalNet-Tiny which rely heavily on the structure, while node and edge features are less useful, expander edges can interfere with the information propagation.

We observed that on molecular datasets,
virtual nodes generally help substantially,
while expander edges don't help much or can even hurt.
The situation for image-based graphs is the opposite,
where especially on Pascal-VOC, expander graphs are highly useful to the model,
but virtual nodes can cause information bottleneck.

Virtual nodes can also make incorporating ideas like the batching mechanism used in GraphSAGE \citep{hamilton2017inductive} more difficult, where having virtual nodes means every batch includes the whole graph. Expander edges can still be used with batching techniques.

\section{Universality of \textsc{Exphormer}} \label{sec:proofs}
In this section, we detail the universal approximation properties of \textsc{Exphormer}-based graph transformers.

The work of \citet{YunBRRK20} showed that for sequences, transformers are universal approximators, i.e., they can approximate any \emph{permutation equivariant} function mapping one sequence to another arbitrarily closely when provided with enough parameters. A function $f: \R^{d\times n} \to \R^{d\times n}$ is said to be permutation equivariant if $f(\mathbf{X}\mathbf{P}) = f(\mathbf{X})\mathbf{P}$, i.e., if permuting the columns of an input $\mathbf{X}\in\R^{d\times n}$ results in the columns of $f(\mathbf{X})$ being permuted the same way.
\begin{theorem}[\citealp{YunBRRK20}]\label{thm:univapprox-perm}
 Let $1\leq p < \infty$ and $\epsilon > 0$. For any function $f: \R^{d\times n} \to \R^{d\times n}$ that is permutation equivariant, there exists a transformer network $g$ such that $\ell^p(f, g) < \epsilon$.
\end{theorem}
The same work shows an extension to all (not necessarily permutation equivariant) sequence-to-sequence functions that are defined on a compact domain, say, $[0,1]^{d\times n}$ provided that one uses a positional encoding. More specifically, for any transformer $g: \R^{d\times n}\to\R^{d\times n}$, one can define a \emph{transformer with positional encoding} $g_p:\R^{d\times n}\to\R^{d\times n}$ such that $g_p(\mathbf{X}) = g(\mathbf{X}+\mathbf{E})$. The following results shows that trainable positional encodings allow a transformer to approximate any sequence-to-sequence continuous function on the compact domain.
\begin{theorem}[\citealp{YunBRRK20}]\label{thm:univapprox-trans}
 Let $1\leq p < \infty$ and $\epsilon > 0$. For any continuous function $f: [0,1]^{d\times n} \to \R^{d\times n}$ that is permutation equivariant, there exists a transformer with positional encoding, $g_P$, such that $\ell^p(f,g) < \epsilon$.
\end{theorem}

Note that the above theorems hold for \emph{full} (dense) transformers. However, under certain conditions about the sparsity pattern, one can obtain similar universality for sparse attention mechanisms~\citep{YunCBRRK20}.

One important consideration is that the aforementioned results hold for functions on \emph{sequences}. Since we are concerned with functions on \emph{graphs}, it is interesting to ask what the implications are for graph transformers.

We follow the approach of \citet{kreuzer2021rethinking}: Given a graph $G$, we can view a node transformer as a function from $\R^{n\times n} \to \R^{n\times n}$ which uses the padded adjacency matrix of $G$ as a positional encoding. Similarly, an edge transformer takes as input a sequence $((i,j), \sigma_{i,j})$ with $i,j \in [n]$ and $i\leq j$ such that $\sigma_{i,j} = 1$ if $i$ and $j$ are connected in $G$ or $\sigma_{i,j} = 0$ otherwise. Any ordering of these vectors corresponds to the same graph. Moreover, we can capture the relevant functions going from $\R^{N(N-1)/2\times 2} \to \R^{N(N-1)/2\times 2}$ with permutation equivariance. Ideally, they can be approximated as closely as desired by suitable transformers on the edge input.

Now, simply observe (see~\citealt{kreuzer2021rethinking}) that one can choose a function (in either the node transformer or edge transformer case) that is (a.) invariant under node permutations and (b.) maps non-isomorphic graphs to distinct values. We would like to apply one of the above thoerems to such a function.

However, we cannot quite apply \Cref{thm:univapprox-perm} or \Cref{thm:univapprox-trans}, as it is specifically for full transformers in which all nodes are pairwise connected in the attention interaction graph.

Therefore, we provide the following theorem, which shows that, under reasonable assumptions, \textsc{Exphormer} is able to provide universal approximation properties using just $O(n)$ edges.

\begin{theorem}\label{thm:univapprox}
 Suppose $H$ is the attention graph of \textsc{Exphormer} (which contains $n$ graph nodes and potentially more virtual nodes), augmented with self loops on all nodes. Suppose $H$ satisfies at least one of the following:
 \begin{enumerate}
     \item $H$ contains at least one node which is connected to all $n$ graph nodes (i.e., at least one virtual node is included).
     \item The underlying expander graph of $H$ contains a Hamiltonian path.
 \end{enumerate}
 Then, it follows that a sparse transformer model, with positional encodings and an attention mechanism following $H$, can universally approximate continuous functions $f:[0,1]^{d\times n} \to \mathbb{R}^{d\times n}$. That is, for any $1<p<\infty$ and $\epsilon > 0$, there exists a sparse transformer network $g$, which uses the attention graph $H$ and some positional encodings, such that $\ell^p(f,g) < \epsilon$.
\end{theorem}
Note that in the above theorem, only one of the enumerated conditions needs to be satisfied; in other words, one does not need to use both global attention and expander graph attention components (see \Cref{sec:attentioncomp}). This is relevant, as for certain datasets, it may be desirable to use one of these attention components but not both (see~\Cref{sec:expander-vs-global}), but universal approximation properties do not depend on this choice.

\begin{proof}[Proof of \Cref{thm:univapprox}]
Let $H$ be the attention graph of a modified sparse \textsc{Exphormer} attention mechanism satisfying the conditions in \Cref{thm:univapprox}.

First, assume that condition (1) is satisfied. This implies that there is a subgraph of $H$ which forms a star graph on the $[n]$ graph nodes (plus, potentially, the virtual node).
In this case, the existence of our desired $g$ is directly implied by the following universal approximation result of \citet{ZaheerGDAAOPRWY20}:
\begin{theorem}[\citealp{ZaheerGDAAOPRWY20}]\label{thm:bigbird-universality}
 Let $1 < p < \infty$ and $\epsilon > 0$. For any graph $H$ on $[n]$ that contains the star graph, we have that if $f\in [0,1]^{n\times d} \to \R^{n\times d}$ is a continuous function, then there exists a sparse transformer network $g$ (with trainable positional encodings) such that $\ell^p(f,g) < \epsilon$.
\end{theorem}

Otherwise, assume that instead condition (2) holds. Then, we can apply Theorem 1 of \citet{YunCBRRK20}. In particular, we note that since (a.)\ $H$ contains self-loops, (b.)\ $H$ contains a Hamiltonian path, and (c.)\ $\diam(H) = O(\log n)$ due to \Cref{cor:logdiam}, it follows that the assumptions of the theorem of \citet{YunCBRRK20} are satisfied. (Note that in our setup, $\rho$ is the softmax function, which satisfies their Assumption 2.) Hence, the desired $g$ exists for any $f,p,\varepsilon$.

This completes the proof.
\end{proof}

This result implies that there exist sparse transformers based on the variant of \textsc{Exphormer} in \Cref{thm:univapprox} which can solve graph isomorphism problems. This does not, however, imply the existence of an efficient algorithm for solving graph isomorphism problems, as we have not shown these networks can be efficiently identified; see further discussion by \citet{kreuzer2021rethinking}.

\end{document}